\documentclass[sigconf]{acmart}
\usepackage{algorithm}
\usepackage{algorithmic}
\usepackage{multirow}
\usepackage{tikz}
\usepackage{pgfplots}
\usepackage{colortbl} 
\AtBeginDocument{%
  }

\setcopyright{acmlicensed}
 \copyrightyear{2024} 
\acmYear{2024} 
\setcopyright{acmlicensed}\acmConference[MM '24]{Proceedings of the 32nd
 ACM International Conference on Multimedia}{October 28-November 1,
 2024}{Melbourne, VIC, Australia}
 \acmBooktitle{Proceedings of the 32nd ACM International Conference on
 Multimedia (MM '24), October 28-November 1, 2024, Melbourne, VIC, Australia}
 \acmDOI{10.1145/3664647.3680661}
 \acmISBN{979-8-4007-0686-8/24/10}

\begin{document}

\title{P-RAG: Progressive Retrieval Augmented Generation For Planning on Embodied Everyday Task}

\author{Weiye Xu}
\email{ustcxwy0271@mail.ustc.edu.cn}
\orcid{0009-0005-2302-8721}
\affiliation{%
  \institution{University of Science and Technology of China, China}  
  \city{Hefei}
  \country{China}
}
\author{Min Wang}
\authornote{Corresponding author.}
\email{wangmin@iai.ustc.edu.cn}
\orcid{0000-0003-3048-6980}
\affiliation{%
  \institution{Institute of Artificial Intelligence, Hefei Comprehensive National Science Center}
  \city{Hefei}
  \country{China}
}

\author{Wengang Zhou}
\authornotemark[1]
\email{zhwg@ustc.edu.cn}
\orcid{0000-0003-1690-9836}
\affiliation{%
  \institution{University of Science and Technology of China}
  \city{Hefei}
  \country{China}
}

\author{Houqiang Li}
\email{lihq@ustc.edu.cn}
\orcid{0000-0003-2188-3028}
\affiliation{%
  \institution{University of Science and Technology of China}
  \city{Hefei}
  \country{China}
}

\renewcommand{\shortauthors}{Weiye Xu, Min Wang, Wengang Zhou, \& Houqiang Li}

\begin{abstract}
Embodied Everyday Task is a popular task in the embodied AI community, requiring agents to make a sequence of actions based on natural language instructions and visual observations. Traditional learning-based approaches face two challenges. Firstly, natural language instructions often lack explicit task planning. Secondly, extensive training is required to equip models with knowledge of the task environment. Previous works based on Large Language Model (LLM) either suffer from poor performance due to the lack of task-specific knowledge or rely on ground truth as few-shot samples. To address the above limitations, we propose a novel approach called Progressive Retrieval Augmented Generation (P-RAG), which not only effectively leverages the powerful language processing capabilities of LLMs but also progressively accumulates task-specific knowledge without ground-truth. Compared to the conventional RAG methods, which retrieve relevant information from the database in a one-shot manner to assist generation, P-RAG introduces an iterative approach to progressively update the database. In each iteration, P-RAG retrieves the latest database and obtains historical information from the previous interaction as experiential references for the current interaction. Moreover, we also introduce a more granular retrieval scheme that not only retrieves similar tasks but also incorporates retrieval of similar situations to provide more valuable reference experiences. Extensive experiments reveal that P-RAG achieves competitive results without utilizing ground truth and can even further improve performance through self-iterations.
\end{abstract}





\maketitle

\section{Introduction}
\begin{figure}
  \centering
  \includegraphics[width=\linewidth]{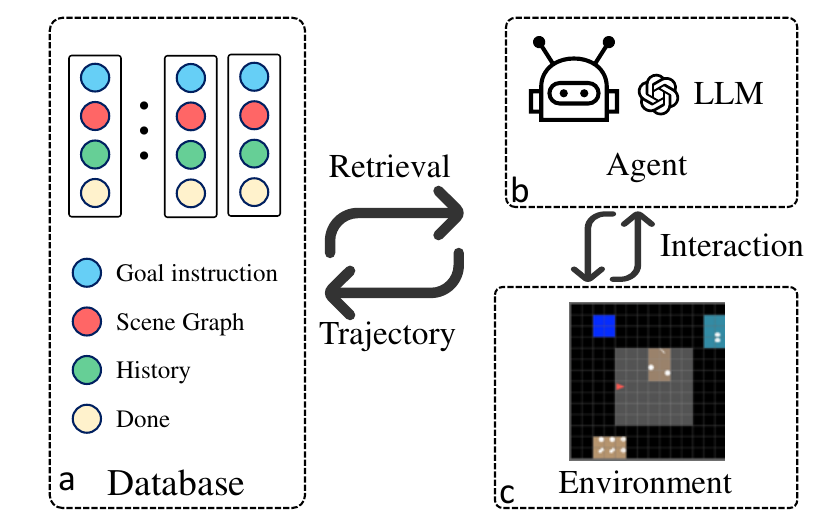}
  \caption{The framework of our Progressive  Retrieval Augmented Generation method. a) The database consists of a list of tuples, each including  goal instruction, scene graph, trajectory history, and whether the task is completed. b) The agent with LLM. c) The interactive
 environment (MINI-BEHAVIOR or ALFRED). The database will update after each complete interaction between the agent and the environment, equipping the agent with increasingly high-quality experiences. }
    \label{fig:fig1}
\end{figure}
Recent years have witnessed the rapid development of Embodied AI (EAI) 
which aims to endow AI agents with the ability to interact with the physical world. Two famous robots in this field from Figure AI and Voxposer \cite{pmlr-v229-huang23b} are both concentrated on Embodied Everyday Task. Due to the risks and instability inherent in conducting experiments in real environments, researchers often opt to train and test algorithms in simulation platforms. Many simulators abstract and integrate key components of the embodied everyday task into their environments, such as MINI-BEHAVIOR \cite{jin2023mini} and ALFRED \cite{ALFRED20}. In these platforms, embodied everyday task is set by a natural language goal instruction guiding the agent's objective or offering step-by-step guidance for aligning the robot's actions with linguistic commands. 

Embodied everyday tasks often encounter three main challenges. 1) Dense reward feedback is typically absent, with the environment only signaling task completion after it has been entirely achieved, usually denoted by a reward of either 1 or 0. Tasks like ``Cleaning up the kitchen only'' in MINI-BEHAVIOR not only feature ambiguous goal instructions without clear subtasks, but also present a challenge for agents due to the difficulty in processing the natural language provided. 2) Another challenge for agents is the variable space of the action in the environment, where actions may not be fixed or there may be invalid actions that cannot be executed in current environment. For example, agents may use cooking or heating to process certain foods, but for everyday household items such as pot plants and shoes, these actions are considered illegal. 3) Certain limitations imposed by real-world circumstances can easily be overlooked. For example, in specific environments, tables may be smaller than usual, unable to accommodate an excessive number of items (in the MINI-BEHAVIOR simulation, each cell is restricted to contain a maximum of three items). However, this information is not commonly known, and even language models trained on textual data may not be aware of such constraints.

Traditional learning-based approaches such as reinforcement learning (RL) can enhance the ability of the model for specific tasks and environments through iterative processes. But they often lack the capability to understand language instructions. The application of large language models (LLM) to embodied everyday task is a highly promising direction. Previous works \cite{vemprala2023chatgpt,hu2023look} have mostly focused on prompt design, which incorporates general commonsense knowledge and addresses the issue of understanding language instructions. Nevertheless, they still lack the ability to possess knowledge specific to particular tasks and environments.

To address the issue of understanding linguistic instructions and imbuing knowledge about specific tasks and situations, we propose a new framework called \textbf{Progressive Retrieval Augmented Generation} (P-RAG) for embodied everyday tasks. It is based on LLM, and designs progressive retrieval to assist in generating actions with specific contextual knowledge iteratively, as shown in Fig.~\ref{fig:fig1}. The progressive  mechanism can iteratively increase the success rate similar to learning-based approach, without involving any training steps. With the text understanding ability of LLM, P-RAG combines both the advantages of both learning-based approaches and pretrained LLM for planning approaches. Indeed, previous works like LLM-planner \cite{song2023llm} have also employed retrieval augmented generation with ground few-shot to enhance the agent's knowledge of specific environments and situations. In comparison, P-RAG has improvements in the following aspects: 1) Instead of using ground-truth action list as few-shot samples, P-RAG utilizes data generated through straightforward interactions with the environment, which is more general for real scenes. 2) In the query, we not only consider searching for trajectory information related to similar tasks but also take into account trajectory information corresponding to similar situations, which provides more valuable context for LLM.

We validate the effectiveness of P-RAG in planning with extensive experiments for everyday tasks. P-RAG outperforms existing methods in few-shot setting, and it provides an effective approach that can further enhance the online planning performance for embodied everyday tasks. Additionally, P-RAG showcases its ability to generalize across different tasks, enabling it to effectively operate in various planning task. Our contributions can be summarized into the following three points.
1) We introduce a new framework for planning with LLM in embodied everyday task. This framework combines the advantages of LLM's prior knowledge and language processing capabilities enhancing the efficiency of utilizing interaction data.
2) Instead of relying on ground truth actions as previous methods do, P-RAG enhances its performance solely through historical trajectories obtained from  interaction of last round.
3) P-RAG outperforms existing methods in utilizing few-shot training datasets, and even provides a self-iteration approach that further enhances performance of testing tasks.

\begin{figure*}[t]
  \centering
  \vspace{-0.2cm}
  \includegraphics[width=0.9\linewidth]{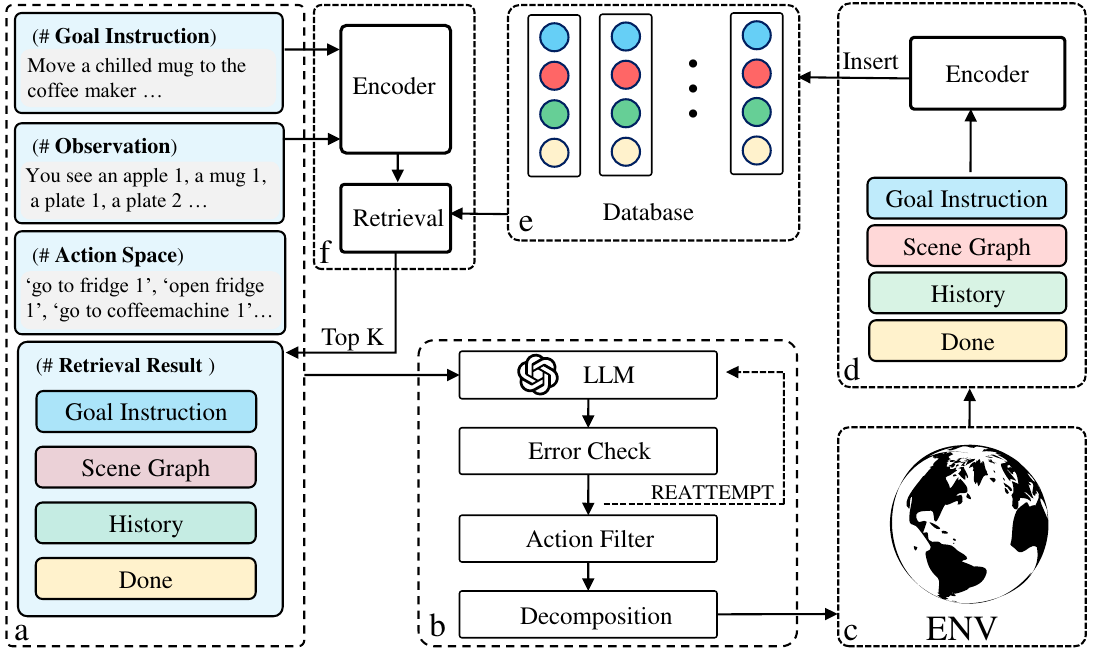}
  \caption{The pipeline of P-RAG in each iteration. ``\#'' stands for the form of text. a) The information transmits to the agent consists of the following four parts: natural language goal instruction, observations obtained from the environment, the action space of the agent, and the retrieval results from the database. b) The agent adopts an LLM to plan a series of actions according to the information in (a). If the LLM produces unsatisfactory content, the agent will initiate a reattempt; otherwise, it will utilize a filtering mechanism to extract the requisite actions from the fields. c) The environment receives actions from the agent and returns observations, along with a ``done" state denoting whether the task is completed. d) Following the completion of each iteration comprising multiple tasks, the database undergoes an update procedure. During each update, it stores the embedding vector of the goal instruction and the scene graph obtained through observation. e) The database contains the trajectories of previous iterations. f) The interface between the database and the agent's information involves two main components. Firstly, the current goal instruction and observation of agent are embedded into vectors, which are further used as query in retrieval augmented process. Secondly, the similarity between query and each database item is computed, and the top K relevant database items are returned to agent.  }
    \label{fig:fig2}
    \vspace{-0.45cm}
\end{figure*}
\section{Related Work}
\subsection{Embodied Everyday Task}
Embodied Everyday Task aims to establish human-centered AI \cite{li2024behavior} that serves human needs, goals, and values by simulating daily human tasks like navigation and manipulation. Due to the increasing demand for human-computer interaction, tasks oriented towards ordinary users are becoming more popular, giving rise to some new tasks such as language-conditional navigation and manipulation \cite{li2024behavior,skirzynski2020language,cui2023no,chen2023bvln,zhuang2022lvln,zhao2022tvln}. These tasks typically involve describing a desired final state using language, requiring algorithms to plan and decompose tasks. The execution process also involves a combination of navigation and manipulation operations. Recently, mainstream approaches to manipulation tasks can roughly be divided into three categories as following: 1) learning-based approaches which needs to train agents in  simulation environments \cite{nair2022r3m, yamada2021motion, zhan2021framework}. 2) LLM-based approaches which  leverage the vast knowledge embedded within large language models \cite{liu2024enhancing,jin2024robotgpt,hu2023look}. 3) hybrid methods \cite{sharan2024plan} combined with the above two approaches where LLMs facilitate task decomposition into subtasks, while meta-skills are honed through training to optimize subtask execution strategies. 

\subsection{Retrieval Augmented Generation}
Retrieval-augmented generation (RAG) is an effective tool for reducing hallucination in large language models like GPT-4 \cite{achiam2023gpt} and improving their performance in generating authentic content. RAG first translates the task description into a query for database retrieval. After obtaining the query results, it integrates them with a prompt for the language model, which then generates the final output. Initially introduced by \cite{NEURIPS2020_6b493230}, this primitive direct approach later became known as Naive RAG (Retrieval-Augmented Generation). It is widely used in various NLP tasks, such as narrative generation \cite{Yamazaki2023narratives}, abstract generation \cite{chen2022target, chen2023topic, cheng2023towards}, and code generation \cite{zhang2023syntax, yu2022bashexplainer}.

Retrieval Augmented Generation for planning represents a promising direction. However, only a limited amount of work has been conducted in this area. Previous methods like LLM-Planner \cite{song2023llm} and RAP \cite{zare2024rap} have utilized the Retrieval Augmented Generation (RAG) approach for planning. However, in their methodologies, ground-truth planning instruction is employed during the training phase or is either utilized during the few-shot learning phase. There has not yet devised a unified approach for interactive tasks without ground truth guidance, which is a more common setting of interactive task like MINI-BEHAVIOR \cite{jin2023mini}.

\subsection{LLM for Planning}
\begin{figure*}[t]
  \centering
  \includegraphics[width=\linewidth]{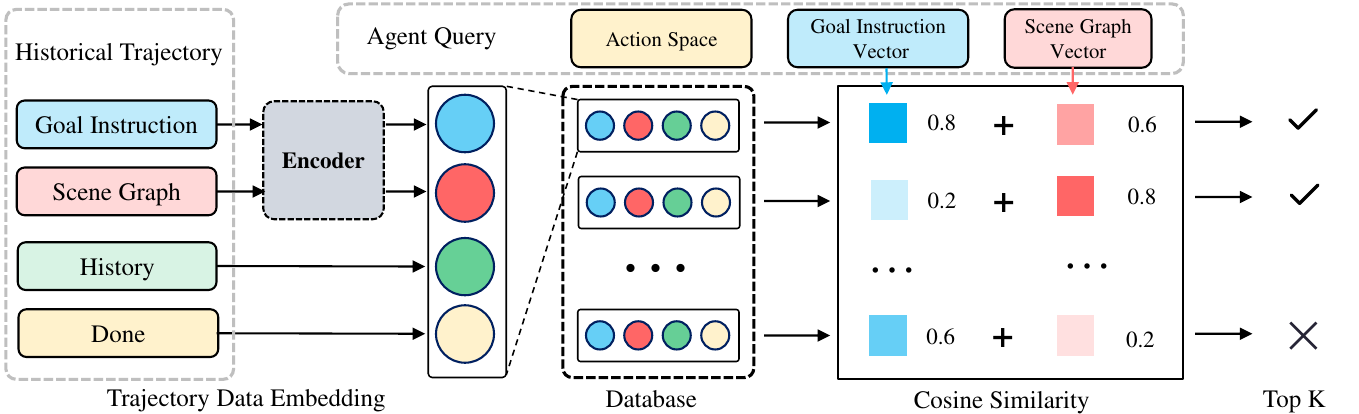}
  \caption{Database Construction and Retrieval. In P-RAG, both the construction of database and retrieval utilize encoding. 1) During the insertion process, four components are inputted: goal instruction, scene graph, history, and done. Among these, goal instruction and scene graph need to undergo sentence embedding to be stored as vectors in the database. 2) When retrieval is required, the current task's goal instruction and scene graph are used as queries. They are also encoded to sentence embedding, while simultaneously computing the similarity score between them and the corresponding vectors in the database. The top K historical trajectory information is then returned based on the aggregated similarity scores.}
    \label{fig:fig3}
\end{figure*}

Large language models like GPT-3.5 and GPT-4 \cite{achiam2023gpt} have significantly advanced various applications. Unlike common text-only tasks, robotic tasks require more prior knowledge and an understanding of interactions with the physical world \cite{lu2022neuro, singh2023progprompt, huang2023visual, huang2022inner}. Large language models (LLMs) are commonly employed for high-level agent planning \cite{vemprala2023chatgpt}. LLMs applied in planning exhibit distinct two advantages \cite{wang2024survey} as following. Firstly, pre-trained on textual data, LLMs inherently incorporate substantial prior knowledge of the physical world, enabling them to understand common robotic tasks. Secondly, LLMs excel at handling tasks involving language instructions, which are prevalent in popular applications.

However, LLMs for planning also encounter several challenges: 1) LLMs possess only general knowledge and lack task-specific information concerning specific environments and tasks. 2) the robustness of LLMs' action outputs is inadequate, sometimes failing to adhere precisely to the required format. 3) LLMs lack comprehensive modeling of the physical world, resulting in decisions that may contravene physical constraints.
\section{Method}
\subsection{Overview}
To accomplish embodied everyday task, we propose a new progressive retrieval augmented planning framework named P-RAG, which integrates better task-specific knowledge progressively into the prompt of large language models (LLMs) for planning. The framework of P-RAG is shown in Fig.~\ref{fig:fig1}. During each interaction episode, the trajectory of agent is collected to iteratively update a dynamic database, which is retrieved to provide task-relevant knowledge from the previous completed interaction. 

Specifically, the detailed pipeline in each episode is shown in Fig.~\ref{fig:fig2}. First, the agent is provided with four types of information including natural language instruction, observations, the action space, and the retrieval results from database. Then, this information is input to LLM for planning, which outputs a series of actions for solving the embodied every task. After that, the environment responds to the receiving actions and converts to the new state, which provides agent new observations and reward. These new trajectory information for agent will be used to update the database, which stores the timely agent experience. In the next interaction episode, this experiential knowledge in the database can be retrieved based on agent information, returning task-relevant knowledge to the LLM for better planning. This pipeline can be iteratively improved, progressively enhancing planning ability by incorporating more successful trajectory information. The detailed components of this pipeline will be elaborated in the following subsections.

\subsection{Agent Input Information}\label{section:Agent Input Information}
Four parts of information will be provided to the agent: goal instructions, observations, action space, and retrieval results. Each part of the information will ultimately be transformed into text format and then integrated into coherent prompt delivered to agent. 

Initially, during the interaction with the task environment, the current task's goal instruction will be provided, such as ``Move a chilled mug to the coffee maker''. During the interaction between P-RAG and the environment, before the agent provides an action, it will first return an observation image to depict the current observed state. We convert the observation image into a scene graph format, which is easier for LLMs to process. We extract the instance labels of all objects from the image and detect the relative relationships between each object, such as ``on the top of'', ``inside of'', and so on. In detail, we slightly adjust the extraction of the scene graph for different interaction environments. For ALFRED, we utilize tools from ALFWORLD, which include interactive TextWorld environments \cite{cote2019textworld} that mirror embodied worlds in the ALFRED dataset \cite{ALFRED20}, to extract instance-level labels of objects observed in the image directly in front of the robot in the environment. After obtaining the instance labels, we first identify the landmarks among them and designate them as key nodes in the scene graph. Next, we organize these labels into a list format, with the key node being placed as the first node in the list for subsequent encoding steps. For the MINI-BEHAVIOR environment, which utilizes gym-minigrid, we sequentially compare the relative relationships between each object by function API in the environment to obtain a relationship matrix with form of each cell in matrix like ``object\_1/object\_2/relationship: True/False''. After obtaining the relationship matrix, to reduce the length of the context input to the LLM, P-RAG retains only those pairs that are true. The action space is also a crucial component, and the action space provided to the LLM consists of high-level actions. Subsequent steps in section \ref{section:Planning with LLM} will involve instruction decomposition, where high-level instructions are converted into low-level actions to be provided to the environment. 

Another component involves historical trajectory information retrieved from database to provide the agent with references, thereby increasing its task-specific knowledge. The retrieval result includes goal instructions, scene graphs, history, and the done state from the previous interaction history. Among them, the goal instruction and scene graph are in the same format as mentioned above.  ``History'' refers to the sequence of $(A_t, O_t)$ pairs, providing the agent with entire pairs in the task of previous round. The ``done'' states indicate whether the task corresponding to these historical pieces of information was completed by the agent at that time. 

\subsection{Planning with LLM}\label{section:Planning with LLM}

The four components of information mentioned above in section \ref{section:Agent Input Information} will be integrated into a single coherent prompt and provided as input to the LLM. Fig.~\ref{fig:fig2} part b illustrates the main workflow of this section, including the LLM, error check, action filter, and decomposition components.
For the first component, we choose GPT-4 and GPT-3.5, renowned for their widespread utilization across various generation tasks, to serve as LLM in P-RAG. The second part involves error checking, where the input consists of the text output from the LLM. It employs regular expression matching and compares the generated actions with those in the action space to ensure adherence to fixed formatting and validity. In the case of invalid actions or non-compliant formatting, a new requirement will be raised for the LLM. Otherwise, the output from the LLM will be passed on to the next component. The third component is the action filter, which extracts the necessary strings from the text generated by the LLM using regular expression matching, providing high-level actions. The final component is decomposition, which translates the high-level actions obtained from the previous component into low-level actions to be executed in the environment. For example, the ``navigation'' high-level action is decomposed using the Fast Marching Method (FMM) \cite{sethian1999fast}. 

Specifically, FMM is a numerical technique designed to tackle boundary value problems arising from the Eikonal equation \cite{sethian1999fast}. We employ a simplified and customized FMM as the tool for decomposing the navigation component from high-level action to low-level action in our process. The FMM process used in P-RAG can be summarized as follows: Firstly, we construct a flat model of the environment with different navigable positions. In MINI-BEHAVIOR, we determine whether each cell in the observation can be covered by the agent. Secondly, we set the current point as the source point and the destination point as the sink point. Then, we compute the Euclidean distance from each point to the source point. Finally, starting from the sink point, we search for the point with the minimum distance value within its one-step neighborhood to serve as the next subgoal point.

\subsection{Database Construction and Retrieval}

P-RAG interacts with the environment through agents to construct and update a historical trajectory database. It retrieves information from this database to guide the agent. The process iterates over a set of tasks across multiple rounds. The first round focuses on constructing the database. In each subsequent round, P-RAG uses retrieval to enhance action generation capabilities and updates the database based on the interaction history of each round.

The database is initialized as an empty collection. The agent interacts with the environment exclusively through goal instructions and observation obtained from each step. After the completion of all task interactions in the first iteration, a round of interaction between the agent and the environment yields valuable information. This information increases the agent's knowledge of specific tasks. The database will store this information and update itself with the latest historical trajectory data after each subsequent iteration.

During each round of data update, the database will retain four key pieces of information for each task interaction sequence: the goal instruction, scene graph, historical data, and completion state. The details of database construction and retrieval pipeline is illustrated in Fig.~\ref{fig:fig3}. When an agent interacts with the environment during a task, it first receives the environment's goal instruction $I_g$ and observation $O_t$. Then it encodes with MiniLM \cite{wang2020minilm} both of them with formula as
\begin{equation}
    Q_{goal} = Encode(I_g) ,
\end{equation}
\begin{equation}
    Q_{obs,t} = Encode(\mathcal{F}(O_t)),
\end{equation}
where $Q_{goal}$ and $Q_{obs,t}$ represent the query embedding vector encoded from goal instruction $I_g$ and observation $O_t$ at time t. $\mathcal{F}$ means scene graph extraction module. With these embedding representation of goal instruction and observation, we use cosine similarity to retrieval the most similar task trajectory in task name or in situation. Each set of embedding vectors serves as keys in the historical records, and will undergo similarity computation with the query. The similarity score for the $n$-th interaction in a task dialogue $s_t$ is calculated as following:
\begin{equation}
    s_n = \text{sim}(Q_{goal},K_{goal}) + \max_{t\in [1,N]}\text{sim} (Q_{obs,n}, K_{obs,t}),
    \label{equ:equ1}
\end{equation}
where $Q_{goal}$ and $K_{goal}$ denote the embedding vectors corresponding to goal instructions, while $Q_{obs,n}$ and $K_{obs,t}$ represent the embedding vectors of scene graph for the $n$-th interaction and the $t$-th interaction, respectively. We select the maximal similarity scores in the $t$-interaction for scene graphs. Ultimately, within the database, P-RAG selects the top K entries as output based on composite score that incorporates both task and scene graph similarities.

\subsection{Progressive Iteration}
\begin{algorithm}[t]
	\caption{P-RAG for Planning} 
        \small
	\label{alg:alg1}
	\begin{algorithmic}
            \REQUIRE{$ENV_{i} ,i=1,2,...,N_{env}$}
            \ENSURE{${A_{i,t}}_{t=1,2,...,N}^{i = 1,2,...,N_{env}}$}
		\STATE $DB \gets \varnothing$;
            \STATE $ENV_{i} \gets RESET ,i=1,2,...,N_{env}$;
            \FOR{$iter in [1,ITER]$}
                \FOR{t in [1,N]}
                    \item $O_{i,t} ,I_{i,goal} \gets ENV_{i}(A_{i,t}),i=0,1,2,... N_{env}$;
                    \item $Q_{obs,i,t} \gets Encode(\mathcal{F}(O_{i,t}))$;
                    \item $Q_{ins,i} \gets Encode( I_{i,goal})$;
                    \IF{$DB \neq  \varnothing$}
                        \item $K_{obs,i,t} , K_{ins,i} \gets DB ,i=1,2,...N_{env}$;
                        \item $s_{i} \gets equation$ \ref{equ:equ1}; 
                        \item $I_{goal},SG,H,D\gets TOPK(DB,\{s_i\}_{i=1}^{N_{env}})$;
                        \item $A_{i,t} \gets Agent(LLM,I_{goal},SG,H,D)$;
                    \ELSE
                        \item $A_{i,t} \gets Agent(LLM)$;
                    \ENDIF
                \ENDFOR
            \ENDFOR
	\end{algorithmic} 
\end{algorithm}

As depicted in Algorithm \ref{alg:alg1}, P-RAG initializes an empty database and initiates successive iterations. At this stage, P-RAG makes decisions solely based on observations of the environment and the generic prior knowledge of LLM. After the first iteration, the historical information from the previous iteration is collected and stored in the database. The database is progressively updated when the agent trajectories of new round are completed. In each subsequent iteration, P-RAG utilizes the collaborative similarity retrieval of scene graph and task name to identify similar tasks and scenes across different tasks. It then provides this information to LLM to facilitate more informed decision-making. The progressive approach is widely employed in learning-based solutions, yet its application in the RAG (Retrieval-Augmented Generation) framework for planning remains a novel paradigm. In scenarios where ground truth is not available, the progressive approach offers a mode of gradual performance improvement compared to the direct method. Moreover, it allows for the accumulation of environment-specific knowledge for the agent through historical trajectories.

\section{Experiment}
\begin{figure*}[th!]
  \centering
  \includegraphics[width=0.95\linewidth]{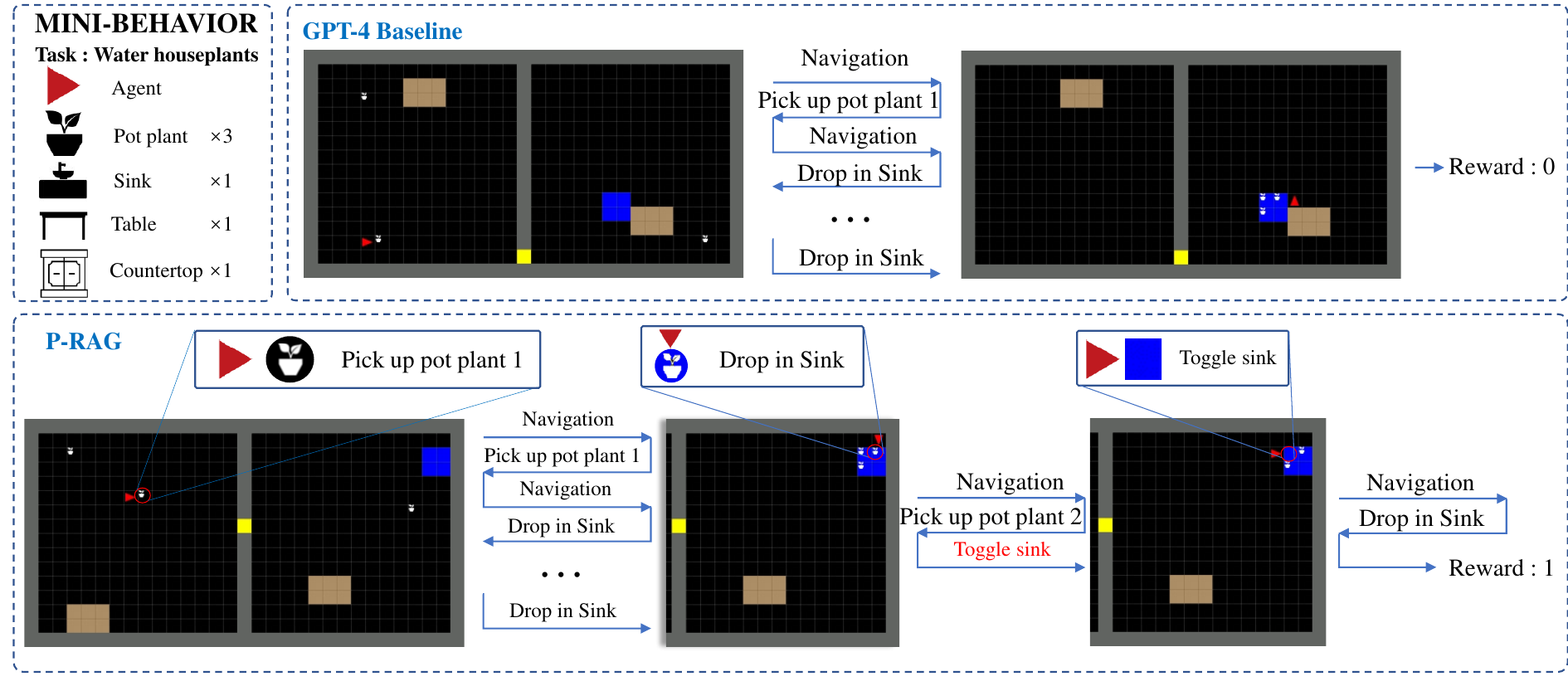}
  \vspace{-0.3cm}
  \caption{Comparison on planning trajectories between GPT-4 baseline and P-RAG. The baseline method follows a decision process of sequentially picking up three pot plants and placing them in the sink, considering the task complete. However, it fails to achieve the task successfully. In contrast, P-RAG utilizes comprehensive historical trajectory information to make decisions, leading to the judgment to toggle the sink and ultimately accomplishing the task. }
    \label{fig:fig4}
    \vspace{-0.4cm}
\end{figure*}
\subsection{Experimental Setup}
\textbf{Dataset}. We select two datasets, MINI-BEHAVIOR \cite{jin2023mini} and ALFRED \cite{ALFRED20}, for our experiments. Extracting 20 activities from BEHAVIOR 1K and abstracting them into a grid environment, the MINI-BEHAVIOR platform offers a comprehensive array of tasks for agents to navigate and manipulate over extended periods. The agent needs to input linguistic instructions and then make decisions to output actions, such as turning left/right, picking up, and so on. Compared to MINI-BEHAVIOR, ALFRED offers more tasks for evaluating the agent and provides realistic observation images. In the ALFRED dataset, we employ the text-world tool from ALFWORLD \cite{shridhar2020alfworld} to convert image-based observations into text format. Since the MINI-BEHAVIOR dataset only consists of 20 environments, we conduct evaluating all of them directly in the experiment. The ALFRED dataset is divided into Valid Seen, Valid Unseen and Train datasets. Following LLM-Planner \cite{song2023llm}, we choose 100 task environments from a pool of 21,023 examples in the ALFRED training set for interaction by P-RAG, which is called Train100 in our experiments. Ensuring a balanced representation, we employ random stratified sampling across all seven task types following the setting to the approach adopted by the LLM-Planner. Valid Seen and Valid Unseen consist of 242 and 85 tasks \cite{ALFRED20}, respectively.

\noindent\textbf{Implementation Details}. Different from the previous works \cite{song2023llm}, we do not require a ground-truth training dataset. All we need is interaction with the environment as the evaluation stage. P-RAG is designed to provide a unified framework for conducting experiments across different datasets. We use the GPT series model as our LLM planner, with the majority of tests utilizing GPT-4 \cite{achiam2023gpt}. In the P-RAG setup, the database will store historical trajectories from previous rounds, with a maximum of 6 iterations. For the ALFRED dataset, we use the JSON format to organize the embedding vector with 384 dimensions of goal instruction and scene graph for each task. The raw trajectory information is recorded in text format in a SQLite database, with an average size of 3.2K bytes per task.

\subsection{Comparison Experiment with the State-of-the-art Methods} 
We first evaluate the ALFRED dataset to compare the success rates (SR) of state-of-the-art methods \cite{zhang2021hierarchical, blukis2022persistent, pashevich2021episodic, song2022one, song2023llm, ahn2022can} with our method. To ensure a fair comparison, after 3 iterations in the training dataset, P-RAG is evaluated on both test datasets. Specifically, we curate 100 task environments from the training set for agent interaction. The database undergoes three iterative updates before subsequent evaluation on the two test datasets.

As illustrated in Table \ref{tab:tab1}, we categorize different methods into two groups based on experimental settings. One group including HiTUT \cite{zhang2021hierarchical}, HLSM \cite{blukis2022persistent}, E.T. \cite{pashevich2021episodic} and  M-TRACK \cite{song2022one} utilizes the full train set with 21,023 instruction and trajectory pairs to derive performance results. Another group including HLSM \cite{blukis2022persistent}, LLM-Planer \cite{song2023llm} and Saycan \cite{ahn2022can} utilizes a smaller subset of training data. Note that within the experimental setup, certain methods denoted with a star (*) utilize step-by-step instructions rather than goal instructions, such as Saycan \cite{ahn2022can} and E.T. \cite{pashevich2021episodic}. The performance comparison in the table reveals that our approach denoted as ``P-RAG (Ours)'' outperforms existing state-of-the-art methods of utilizing few training dataset on both Valid Seen and Valid Unseen dataset. These results demonstrate the effectiveness of our method, which develops the progressive retrieval augmentation to assist the large language model to obtain task-relevant information. Compared to methods trained on the full dataset, P-RAG maintains high performance even under conditions where only approximately 1/200 of the training data is utilized. These results prove that our method possesses better generalizability than these training methods, which usually overfit the training datasets and achieve better performance on the Val Seen dataset than Val Unseen dataset. 

Besides the standard setting of our method, we also construct a self-iteration variant denoted as ``P-RAG (Self-Iter.)''. It constructs progressive iteration over the test dataset, which means the retrieval database is constructed by the trajectories on test dataset. It directly improves the success rates of testing tasks, since the agent obtains better task-relevant experiences about testing tasks by progressive retrieval. From Table 1, with self-iteration on the Valid Unseen dataset, our method denoted as ``P-RAG (Self-Iter.)'' can even outperform all the methods, both with and without using the entire training dataset, which further verifies the effectiveness of the proposed progressive retrieval augmented planning framework.

\begin{table}[t]
    \centering
    \tabcolsep 3pt
    \renewcommand\arraystretch{0.9}
    \captionsetup{width=.48\textwidth, skip=10pt} 
    \begin{tabular}{lcccc}
    \toprule
    \textbf{Model} & \textbf{Dataset} & \textbf{G.T.} &
    \textbf{Valid Unseen} & \textbf{Valid Seen} \\
    \midrule
    HiTUT \cite{zhang2021hierarchical} & Full & \checkmark & 10.23 & 18.41 \\
    HLSM \cite{blukis2022persistent}  & Full & \checkmark  & \textbf {18.28} & 29.63  \\
    *E.T. \cite{pashevich2021episodic}  & Full & \checkmark & 7.32 & \textbf{46.59} \\
    *HiTUT \cite{zhang2021hierarchical}  & Full & \checkmark  & 12.44 & 25.24 \\
    *M-TRACK \cite{song2022one} & Full & \checkmark & 17.29 & 26.70 \\
    \midrule
    *HLSM \cite{blukis2022persistent} & Part & \checkmark & 0.00 & 0.13 \\
    LLM-Planer \cite{song2023llm} & Part & \checkmark  & 12.92   & 13.53   \\
    *Saycan \cite{ahn2022can} & Part & \checkmark & 9.88 & 12.3 \\
    GPT-4 \cite{achiam2023gpt} & - & $\times$ & 7.05 & 17.46 \\
    \textbf{P-RAG (Ours)} & Part & $\times$ & \textbf{14.11} & \textbf{18.2} \\
    \midrule
    \textbf{P-RAG (Self-Iter.)} & - & $\times$ & \textbf{27.4} & 19.05 \\
     \bottomrule
    \end{tabular}
    \caption{Comparison of Different Method on ALFRED. Star (*) stands for using step-by-step instruction instead of goal instruction. G.T. means ground-truth action. In the ``Dataset'' column, ``Full'' indicates the utilization of the entire training dataset, while ``Part'' indicates sampling from a subset of the training dataset. ``P-RAG (Self-Iter.)'' represents the iterative updates of P-RAG on the same group of task datasets.}
    \vspace{-0.7cm}
    \label{tab:tab1}
    \vspace{-0.2cm}
\end{table}

We also evaluate the performance of P-RAG on MINI-BEHAVIOR which lacks ground-truth actions annotation, and therefore can be used to evaluate our method and compared methods in weak supervision setting. As shown in Table \ref{tab:tab2}, we evaluate through three evaluation metrics: Total success rate (SR), which represents the success average by episodes; Task success rate (SR), indicating success average by tasks; Success weighted by Path Length (SPL), evaluated according to the following formula:
\begin{equation}
    SPL = \frac{1}{N} \sum_{i=1}^{N} S_i \frac{L_i}{max(L_i,P_i)},
\end{equation}
where N is total number of evalutional episodes, $S_i$ represents whether the current episode is successful, $L_i$ denotes the number of actions used by the agent, and $P_i$ represents the number of steps needed to complete the task in the shortest possible manner.

We compare the success rate of P-RAG relative to its corresponding baseline with LLM selecting either GPT-4 or GPT-3.5. From the results in the table, we can observe that P-RAG demonstrates significant improvements of 1.7\% and 2.5\% compared to the baselines of GPT-4 and GPT-3.5, respectively. Although simple and lightweight, MINI-BEHAVIOR presents a formidable challenge for popular Reinforcement Learning algorithms, particularly in the absence of a dense reward signal. The vanilla PPO algorithm is only able to achieve a valid success rate (approximately 8\%) after training 1e6 steps and evaluating on a single task (not work in other tasks) within MINI-BEHAVIOR \cite{jin2023mini}. In contrast, P-RAG achieves 16.7\% Total SR requiring iterations of no more than 6 eposides and accomplish 5 tasks (compare to single task of vanilla PPO algorithm), demonstrating its effectiveness across different environments and its few-shot property. We select one group of task trajectories as visualization result, as shown in Fig.~\ref{fig:fig4}. The visualization of trajectory demonstrates the decision-making processes of P-RAG and the GPT-4 baseline for the task ``water houseplants''. From the visualization results, it can be observed that P-RAG is able to make judgments with more task-specific knowledge by referring to historical trajectory information.
\begin{table}[t!]
    \centering
    \tabcolsep 2pt
    \renewcommand\arraystretch{0.9}
    \captionsetup{width=.45\textwidth, skip=10pt} 
    \begin{tabular}{lcccc}
    \toprule
    \textbf{MINI-BEHAVIOR} & \textbf{GPT-4\cite{achiam2023gpt}} & \textbf{P-RAG-4} &
    \textbf{GPT-3.5} & \textbf{P-RAG-3.5} \\
    \midrule
    Total SR  & 15\% & 16.7\% & 7.5\% & 10\% \\
    Task SR  & 20\% & 25\% & 20\% & 20\% \\
    SPL & 13.8\% & 15\% & 7.5\% & 9.5\% \\
    \bottomrule
    \end{tabular}
    \caption{Comparison of Retrieval Augmented Models on MINI-BEHAVIOR. GPT-3.5 and GPT-4 represent the results of each as the baseline LLM planner. P-RAG-3.5 and P-RAG-4 represent the results of setting the LLM in P-RAG as GPT-3.5 and GPT-4, respectively.}
    \label{tab:tab2}
    \vspace{-0.6cm}
\end{table}
\subsection{Ablation Study \& Parameter Analysis}

\textbf{Progressive Iteration}. We conduct additional experiments on P-RAG regarding self-revolution to validate its progressive effect. Compared to the previous approach of iterating on the training dataset, we directly conduct progressive retrieval iteration on the same dataset, which we call self-iteration. In particular, we choose to conduct self-iteration on the Valid Seen dataset and training dataset (where there may be overlap between the Valid Seen and Train datasets), respectively. From Table \ref{tab:tab3}, there is an improvement after progressive iteration in both task datasets. P-RAG not only demonstrates the advantages of performance in standard experimental setting as shown in the first row ``Train100'', but also achieves performance improvement through progressive iteration on the Valid Unseen dataset as shown in the second row in Table \ref{tab:tab3}.

\begin{table}[t]
    \centering
    \tabcolsep 5pt
    \renewcommand\arraystretch{0.9}
    \captionsetup{width=.85\textwidth, skip=10pt} 
    \begin{tabular}{lcccc}
    \toprule
    \textbf{Dataset} & \textbf{Original} & \textbf{1st Iter.} &
    \textbf{2nd Iter.} & \textbf{3rd Iter.} \\
    \midrule
    Train100  & 5\% & 9\% & 10\% & 11\% \\
    Valid Unseen & 7.05\% & 14.11\% & 20.00\% & 22.35\% \\
    \bottomrule
    \end{tabular}
    \caption{ Success Rate with the Iteration Number on ALFRED.}
    \label{tab:tab3}
    \vspace{-0.85cm}
\end{table}

\begin{figure}[ht]
  \centering
  \resizebox{0.32 \textwidth}{!}{
    \begin{tikzpicture}
      \begin{axis}[
          xlabel={Number of Iterations},
          ylabel={Success Rate (\%)},
          legend pos=north west,
          grid=major,
           ylabel near ticks, 
        ]
        \addplot[color=blue, mark=o,thick] coordinates {
          (0, 7.05)
          (1, 14.11)
          (2, 20.00)
          (3, 25.00)
          (4, 26.2)
          (5, 27.4)
          (6, 27.4)
        };
        \addlegendentry{Valid Unseen};
  
        \addplot[color=red, mark=triangle,thick] coordinates {
          (0, 5)
          (1, 9)
          (2, 10)
          (3, 11)
          (4, 10)
        };
        \addlegendentry{Train100};
      \end{axis}
    \end{tikzpicture}
  }
  \vspace{-0.3cm}
  \caption{Performance with Iteration Number on ALFRED.}
  \label{fig:plot1}
  \vspace{-0.3cm}
\end{figure}
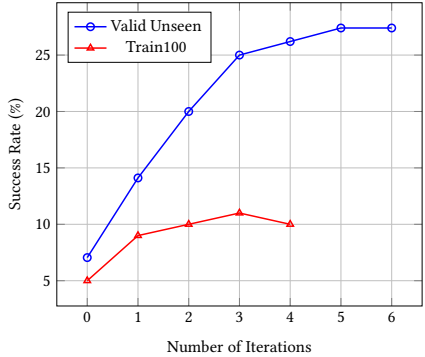

We also conduct in-depth experiments on performance saturation, by enhancing retrieval through updating historical information from the previous round after encoding it into the database. Subsequently, we update the database based on the current round's historical trajectory, iterating multiple times until saturation is achieved, as shown in Fig.~\ref{fig:plot1}. From Fig.~\ref{fig:plot1}, P-RAG achieves significant performance improvement through iteration. In the ALFRED Valid Unseen dataset, the success rate is improved from 7.05\% at the beginning to 27.4\% after 5 rounds of iteration. Similarly, in the ALFRED Train 100 dataset, the success rate increases from 5\% at the beginning to 11\% after 3 rounds of iteration. Both of them all eventually reach performance saturation. From the curves, it can be observed that the curve of success rate increase in each iteration of P-RAG until convergence, indicating that it is approaching the performance limit of LLM as a Planner on the testing tasks.

\noindent\textbf{Correlation Between Iteration}. We also continue to study the correlation between successfully completed tasks during the iterative process of P-RAG. Table \ref{tab:tab4} displays the correlations between tasks completed at different iteration stages. The data values in the table are derived from iterations 0, 1, 2, and 3 in the ALFRED dataset. Each pair of True and False corresponds to successful and failed tasks in a single iteration. The value within each cell indicates the ratio of occurrences of ``done'' compared to the total occurrences of True/False in the corresponding row's iteration. Considering the value located in the top-left corner of the table, this value signifies the proportion of successful tasks from the initial iteration that remain successful after the first subsequent iteration.

\begin{table}[ht]
\centering
\small
\begin{tabular}{lccccccc}
\toprule
& & \multicolumn{2}{c}{\textbf{Next 1 Iter.}} & \multicolumn{2}{c}{\textbf{Next 2 Iter.}} & \multicolumn{2}{c}{\textbf{Next 3 Iter.}} \\
& & \textbf{True} & \textbf{False} & \textbf{True} & \textbf{False} & \textbf{True} & \textbf{False} \\
\midrule
\multirow{2}{*}{\textbf{0th Iter.}}& \textbf{True} & 83.3\% & 16.7\% & 100\% & 0.0\% & 83.3\% & 16.7\% \\
& \textbf{False} & 6.8\% & 93.2\% & 14.9\% & 85.1\% & 17.6\% & 82.4\% \\
\midrule
\multirow{2}{*}{\textbf{1st Iter.}}& \textbf{True} & 72.7\% & 27.3\% & 72.7\% & 27.3\% & - & - \\
& \textbf{False} & 12.3\% & 87.7\% & 13.7\% & 86.3\% & - & - \\
\midrule
\multirow{2}{*}{\textbf{2nd Iter.}}& \textbf{True} & 88.2\% & 11.8\% & - & - & - & - \\
& \textbf{False} & 4.5\% & 95.5\% & - & - & - & - \\
\bottomrule
\end{tabular}

\caption{Relation Between Iteration on ALFRED. ``N-th Iter.'' represents the completion status of tasks after the n-th iteration of P-RAG. For the row labeled ``Next m-th.'' each cell represents the comparison between the (n+m)-th iteration and the n-th iteration. The value in each cell represents the proportion of True/False occurrences of ``done'' in its column compared to the total occurrences of True/False in its corresponding row's iterations.}
\label{tab:tab4}
\vspace{-0.5cm}
\end{table}

By analyzing the data in Table \ref{tab:tab4}, it is evident that P-RAG predominantly achieves success in subsequent iterations based on the success of the previous iteration. Additionally, it demonstrates the capability to succeed in tasks that were unsuccessful in the previous iteration. This highlights how P-RAG not only effectively maintains performance from the previous iteration but also leverages information from unsuccessful trajectories to achieve new successes. This underscores the sources of performance enhancement in P-RAG.

\begin{table}[t]
  \centering
    \renewcommand\arraystretch{0.9}
  \captionsetup{width=.49\textwidth, skip=10pt} 
  \begin{tabular}{cccccc}
    \toprule 
    \multicolumn{3}{c}{\textbf{P-RAG (Self-Iter.)}} & \multirow{2}{*}{\textbf{P-RAG}} & \multirow{2}{*}{\textbf{P-RAG}} & \multirow{2}{*}{\textbf{GPT-4 \cite{achiam2023gpt}}} \\
    \cmidrule(lr){1-3}
    1st  &  2nd & 3rd & & \textbf{w/o SG } & \\
    \midrule 
     14.11\% & 20.00\% & 22.35\% & 14.11\% &  13.10\% & 7.05\% \\
    \bottomrule
  \end{tabular}
  \caption{Ablation with Scene Graph on ALFRED Valid Unseen. The  first 3 columns represent P-RAG with 3 rounds of self-iterations on Valid Unseen dataset.  The following two columns indicate testing on Valid Unseen after iterations on Train100. }
  \label{tab:tab5}
  \vspace{-0.9cm}
\end{table}
    
\noindent\textbf{Ablation Study}. We demonstrate through ablation experiments to verify 1) the effectiveness of the progressive approach and 2) the effectiveness of the joint retrieval of scene graph and task name compared to using task name alone.

For the former, in the last row of Table \ref{tab:tab1}, P-RAG is demonstrated to be effective through iterations on both Valid Unseen and Valid Seen. Through iterations, P-RAG outperforms nearly all the methods, including step-by-step instruction and training on the full dataset, in performance on Valid Unseen without training and with a few number of samples. Similarly, in Table \ref{tab:tab2}, conducting two sets of experiments with LLM as GPT-4 and GPT-3.5 also demonstrates the effectiveness of the progressive method.

To further evaluate the effectiveness of the progressive mechanism, we conduct extra ablation experiments on the ALFRED Valid Unseen dataset, as shown in Table \ref{tab:tab5}. The interpretation of the data in the table is as following. The ``GPT-4'' column represents the results by using the GPT-4 baseline for testing. The ``P-RAG (Self-Iter.)'' colunm represents the results obtained through testing with 1, 2 and 3 iterations, respectively. The ``P-RAG'' column represents the results obtained by iteratively using task name and scene graph joint retrieval on the Train100 dataset and then testing on the Valid Unseen dataset. The ``P-RAG w/o SG'' column represents the results obtained by solely using task name for retrieval, iteratively on the Train100 and then testing on the Valid Unseen dataset. From the table, it is evident that compared to not utilizing scene graph for joint retrieval, P-RAG exhibits significant improvements. This demonstrates the effectiveness of the progressive mechanism and the necessity of incorporating scene graph and task name retrieval.

\section{Conclusion}

We propose a novel progressive retrieval augmented generation framework called P-RAG for planning to address the issue of lacking task-specific and scene-specific knowledge in LLM. P-RAG distinguishes previous methods by eliminating the need for ground truth and achieves promising performance with just a few task interactions. By collaboration with task name and scene graph for P-RAG to retrieval, we ensure historical records selected from database are not only similar in tasks but also in situations. P-RAG's unique iterative capability continuously acquires historical information through interaction. By leveraging enhanced retrieval capabilities, P-RAG incrementally accumulates task-specific knowledge during iterations, thereby improving its planning performance. The experiments demonstrate the effectiveness of P-RAG compared with the state-of-the-art planning methods for Embodied Everyday task. We hope that the ground-truth-free capability in P-RAG can be applied to a broader range of planning tasks and deliver even more outstanding performance.
\begin{acks}
This work was supported by National Natural Science Foundation of China under Contract 62102128 \& 62021001, and the Youth Innovation Promotion Association CAS. It was also supported by GPU cluster built by MCC Lab of Information Science and Technology Institution, USTC, and the Supercomputing Center of the USTC. 
\end{acks}

\bibliographystyle{ACM-Reference-Format}
\bibliography{samples/sample-base}


\begin{thebibliography}{41}


\ifx \showCODEN    \undefined \def \showCODEN     #1{\unskip}     \fi
\ifx \showDOI      \undefined \def \showDOI       #1{#1}\fi
\ifx \showISBNx    \undefined \def \showISBNx     #1{\unskip}     \fi
\ifx \showISBNxiii \undefined \def \showISBNxiii  #1{\unskip}     \fi
\ifx \showISSN     \undefined \def \showISSN      #1{\unskip}     \fi
\ifx \showLCCN     \undefined \def \showLCCN      #1{\unskip}     \fi
\ifx \shownote     \undefined \def \shownote      #1{#1}          \fi
\ifx \showarticletitle \undefined \def \showarticletitle #1{#1}   \fi
\ifx \showURL      \undefined \def \showURL       {\relax}        \fi
\providecommand\bibfield[2]{#2}
\providecommand\bibinfo[2]{#2}
\providecommand\natexlab[1]{#1}
\providecommand\showeprint[2][]{arXiv:#2}

\bibitem[Achiam et~al\mbox{.}(2023)]%
        {achiam2023gpt}
\bibfield{author}{\bibinfo{person}{Josh Achiam}, \bibinfo{person}{Steven Adler}, \bibinfo{person}{Sandhini Agarwal}, \bibinfo{person}{Lama Ahmad}, \bibinfo{person}{Ilge Akkaya}, \bibinfo{person}{Florencia~Leoni Aleman}, \bibinfo{person}{Diogo Almeida}, \bibinfo{person}{Janko Altenschmidt}, \bibinfo{person}{Sam Altman}, \bibinfo{person}{Shyamal Anadkat}, {et~al\mbox{.}}} \bibinfo{year}{2023}\natexlab{}.
\newblock \showarticletitle{GPT-4 technical report}.
\newblock \bibinfo{journal}{\emph{arXiv preprint arXiv:2303.08774}} (\bibinfo{year}{2023}).
\newblock


\bibitem[Ahn et~al\mbox{.}(2022)]%
        {ahn2022can}
\bibfield{author}{\bibinfo{person}{Michael Ahn}, \bibinfo{person}{Anthony Brohan}, \bibinfo{person}{Noah Brown}, \bibinfo{person}{Yevgen Chebotar}, \bibinfo{person}{Omar Cortes}, \bibinfo{person}{Byron David}, \bibinfo{person}{Chelsea Finn}, \bibinfo{person}{Chuyuan Fu}, \bibinfo{person}{Keerthana Gopalakrishnan}, \bibinfo{person}{Karol Hausman}, {et~al\mbox{.}}} \bibinfo{year}{2022}\natexlab{}.
\newblock \showarticletitle{Do as i can, not as i say: Grounding language in robotic affordances}.
\newblock \bibinfo{journal}{\emph{arXiv preprint arXiv:2204.01691}} (\bibinfo{year}{2022}).
\newblock


\bibitem[Blukis et~al\mbox{.}(2022)]%
        {blukis2022persistent}
\bibfield{author}{\bibinfo{person}{Valts Blukis}, \bibinfo{person}{Chris Paxton}, \bibinfo{person}{Dieter Fox}, \bibinfo{person}{Animesh Garg}, {and} \bibinfo{person}{Yoav Artzi}.} \bibinfo{year}{2022}\natexlab{}.
\newblock \showarticletitle{A persistent spatial semantic representation for high-level natural language instruction execution}. In \bibinfo{booktitle}{\emph{Conference on Robot Learning}}. PMLR, \bibinfo{pages}{706--717}.
\newblock


\bibitem[Chen et~al\mbox{.}(2023b)]%
        {chen2023bvln}
\bibfield{author}{\bibinfo{person}{Jingwen Chen}, \bibinfo{person}{Jianjie Luo}, \bibinfo{person}{Yingwei Pan}, \bibinfo{person}{Yehao Li}, \bibinfo{person}{Ting Yao}, \bibinfo{person}{Hongyang Chao}, {and} \bibinfo{person}{Tao Mei}.} \bibinfo{year}{2023}\natexlab{b}.
\newblock \showarticletitle{Boosting Vision-and-Language Navigation with Direction Guiding and Backtracing}.
\newblock \bibinfo{journal}{\emph{ACM Trans. Multimedia Comput. Commun. Appl.}} \bibinfo{volume}{19}, \bibinfo{number}{1}, Article \bibinfo{articleno}{9} (\bibinfo{date}{jan} \bibinfo{year}{2023}), \bibinfo{numpages}{16}~pages.
\newblock
\showISSN{1551-6857}
\urldef\tempurl%
\url{https://doi.org/10.1145/3526024}
\showDOI{\tempurl}


\bibitem[Chen et~al\mbox{.}(2023a)]%
        {chen2023topic}
\bibfield{author}{\bibinfo{person}{Xiuying Chen}, \bibinfo{person}{Mingzhe Li}, \bibinfo{person}{Shen Gao}, \bibinfo{person}{Xin Cheng}, \bibinfo{person}{Qiang Yang}, \bibinfo{person}{Qishen Zhang}, \bibinfo{person}{Xin Gao}, {and} \bibinfo{person}{Xiangliang Zhang}.} \bibinfo{year}{2023}\natexlab{a}.
\newblock \showarticletitle{A topic-aware summarization framework with different modal side information}.
\newblock \bibinfo{journal}{\emph{International ACM SIGIR Conference on Research and Development in Information Retrieval}} (\bibinfo{year}{2023}), \bibinfo{pages}{1416--1425}.
\newblock


\bibitem[Cheng et~al\mbox{.}(2023)]%
        {cheng2023towards}
\bibfield{author}{\bibinfo{person}{Xin Cheng}, \bibinfo{person}{Shen Gao}, \bibinfo{person}{Yuchi Zhang}, \bibinfo{person}{Yongliang Wang}, \bibinfo{person}{Xiuying Chen}, \bibinfo{person}{Mingzhe Li}, \bibinfo{person}{Dongyan Zhao}, {and} \bibinfo{person}{Rui Yan}.} \bibinfo{year}{2023}\natexlab{}.
\newblock \showarticletitle{Towards personalized review summarization by modeling historical reviews from customer and product separately}.
\newblock \bibinfo{journal}{\emph{arXiv preprint arXiv:2301.11682}} (\bibinfo{year}{2023}).
\newblock


\bibitem[C{\^o}t{\'e} et~al\mbox{.}(2019)]%
        {cote2019textworld}
\bibfield{author}{\bibinfo{person}{Marc-Alexandre C{\^o}t{\'e}}, \bibinfo{person}{Akos K{\'a}d{\'a}r}, \bibinfo{person}{Xingdi Yuan}, \bibinfo{person}{Ben Kybartas}, \bibinfo{person}{Tavian Barnes}, \bibinfo{person}{Emery Fine}, \bibinfo{person}{James Moore}, \bibinfo{person}{Matthew Hausknecht}, \bibinfo{person}{Layla El~Asri}, \bibinfo{person}{Mahmoud Adada}, {et~al\mbox{.}}} \bibinfo{year}{2019}\natexlab{}.
\newblock \showarticletitle{Textworld: A learning environment for text-based games}. In \bibinfo{booktitle}{\emph{Computer Games: Workshop International Conference on Artificial Intelligence}}. Springer, \bibinfo{pages}{41--75}.
\newblock


\bibitem[Cui et~al\mbox{.}(2023)]%
        {cui2023no}
\bibfield{author}{\bibinfo{person}{Yuchen Cui}, \bibinfo{person}{Siddharth Karamcheti}, \bibinfo{person}{Raj Palleti}, \bibinfo{person}{Nidhya Shivakumar}, \bibinfo{person}{Percy Liang}, {and} \bibinfo{person}{Dorsa Sadigh}.} \bibinfo{year}{2023}\natexlab{}.
\newblock \showarticletitle{No, to the right: Online language corrections for robotic manipulation via shared autonomy}. In \bibinfo{booktitle}{\emph{ACM/IEEE International Conference on Human-Robot Interaction}}. \bibinfo{pages}{93--101}.
\newblock


\bibitem[Hu et~al\mbox{.}(2023)]%
        {hu2023look}
\bibfield{author}{\bibinfo{person}{Yingdong Hu}, \bibinfo{person}{Fanqi Lin}, \bibinfo{person}{Tong Zhang}, \bibinfo{person}{Li Yi}, {and} \bibinfo{person}{Yang Gao}.} \bibinfo{year}{2023}\natexlab{}.
\newblock \showarticletitle{Look before you leap: Unveiling the power of gpt-4v in robotic vision-language planning}.
\newblock \bibinfo{journal}{\emph{arXiv preprint arXiv:2311.17842}} (\bibinfo{year}{2023}).
\newblock


\bibitem[Huang et~al\mbox{.}(2023a)]%
        {huang2023visual}
\bibfield{author}{\bibinfo{person}{Chenguang Huang}, \bibinfo{person}{Oier Mees}, \bibinfo{person}{Andy Zeng}, {and} \bibinfo{person}{Wolfram Burgard}.} \bibinfo{year}{2023}\natexlab{a}.
\newblock \showarticletitle{Visual language maps for robot navigation}. In \bibinfo{booktitle}{\emph{IEEE International Conference on Robotics and Automation}}. IEEE, \bibinfo{pages}{10608--10615}.
\newblock


\bibitem[Huang et~al\mbox{.}(2023b)]%
        {pmlr-v229-huang23b}
\bibfield{author}{\bibinfo{person}{Wenlong Huang}, \bibinfo{person}{Chen Wang}, \bibinfo{person}{Ruohan Zhang}, \bibinfo{person}{Yunzhu Li}, \bibinfo{person}{Jiajun Wu}, {and} \bibinfo{person}{Li Fei-Fei}.} \bibinfo{year}{2023}\natexlab{b}.
\newblock \showarticletitle{VoxPoser: Composable 3D value maps for robotic manipulation with language models}. In \bibinfo{booktitle}{\emph{Conference on Robot Learning}}, Vol.~\bibinfo{volume}{229}. \bibinfo{publisher}{PMLR}, \bibinfo{pages}{540--562}.
\newblock


\bibitem[Huang et~al\mbox{.}(2022)]%
        {huang2022inner}
\bibfield{author}{\bibinfo{person}{Wenlong Huang}, \bibinfo{person}{Fei Xia}, \bibinfo{person}{Ted Xiao}, \bibinfo{person}{Harris Chan}, \bibinfo{person}{Jacky Liang}, \bibinfo{person}{Pete Florence}, \bibinfo{person}{Andy Zeng}, \bibinfo{person}{Jonathan Tompson}, \bibinfo{person}{Igor Mordatch}, \bibinfo{person}{Yevgen Chebotar}, {et~al\mbox{.}}} \bibinfo{year}{2022}\natexlab{}.
\newblock \showarticletitle{Inner monologue: Embodied reasoning through planning with language models}.
\newblock \bibinfo{journal}{\emph{arXiv preprint arXiv:2207.05608}} (\bibinfo{year}{2022}).
\newblock


\bibitem[Jin et~al\mbox{.}(2023)]%
        {jin2023mini}
\bibfield{author}{\bibinfo{person}{Emily Jin}, \bibinfo{person}{Jiaheng Hu}, \bibinfo{person}{Zhuoyi Huang}, \bibinfo{person}{Ruohan Zhang}, \bibinfo{person}{Jiajun Wu}, \bibinfo{person}{Li Fei-Fei}, {and} \bibinfo{person}{Roberto Mart{\'\i}n-Mart{\'\i}n}.} \bibinfo{year}{2023}\natexlab{}.
\newblock \showarticletitle{Mini-BEHAVIOR: A Procedurally generated benchmark for long-horizon decision-making in embodied AI}.
\newblock \bibinfo{journal}{\emph{arXiv preprint arXiv:2310.01824}} (\bibinfo{year}{2023}).
\newblock


\bibitem[Jin et~al\mbox{.}(2024)]%
        {jin2024robotgpt}
\bibfield{author}{\bibinfo{person}{Yixiang Jin}, \bibinfo{person}{Dingzhe Li}, \bibinfo{person}{A Yong}, \bibinfo{person}{Jun Shi}, \bibinfo{person}{Peng Hao}, \bibinfo{person}{Fuchun Sun}, \bibinfo{person}{Jianwei Zhang}, {and} \bibinfo{person}{Bin Fang}.} \bibinfo{year}{2024}\natexlab{}.
\newblock \showarticletitle{Robotgpt: Robot manipulation learning from chatgpt}.
\newblock \bibinfo{journal}{\emph{IEEE Robotics and Automation Letters}} (\bibinfo{year}{2024}).
\newblock


\bibitem[Lewis et~al\mbox{.}(2020)]%
        {NEURIPS2020_6b493230}
\bibfield{author}{\bibinfo{person}{Patrick Lewis}, \bibinfo{person}{Ethan Perez}, \bibinfo{person}{Aleksandra Piktus}, \bibinfo{person}{Fabio Petroni}, \bibinfo{person}{Vladimir Karpukhin}, \bibinfo{person}{Naman Goyal}, \bibinfo{person}{Heinrich K\"{u}ttler}, \bibinfo{person}{Mike Lewis}, \bibinfo{person}{Wen-tau Yih}, \bibinfo{person}{Tim Rockt\"{a}schel}, \bibinfo{person}{Sebastian Riedel}, {and} \bibinfo{person}{Douwe Kiela}.} \bibinfo{year}{2020}\natexlab{}.
\newblock \showarticletitle{Retrieval-Augmented generation for knowledge-intensive NLP Tasks}. In \bibinfo{booktitle}{\emph{Advances in Neural Information Processing Systems}}, Vol.~\bibinfo{volume}{33}. \bibinfo{publisher}{Curran Associates, Inc.}, \bibinfo{pages}{9459--9474}.
\newblock


\bibitem[Li et~al\mbox{.}(2024)]%
        {li2024behavior}
\bibfield{author}{\bibinfo{person}{Chengshu Li}, \bibinfo{person}{Ruohan Zhang}, \bibinfo{person}{Josiah Wong}, \bibinfo{person}{Cem Gokmen}, \bibinfo{person}{Sanjana Srivastava}, \bibinfo{person}{Roberto Mart{\'\i}n-Mart{\'\i}n}, \bibinfo{person}{Chen Wang}, \bibinfo{person}{Gabrael Levine}, \bibinfo{person}{Wensi Ai}, \bibinfo{person}{Benjamin Martinez}, {et~al\mbox{.}}} \bibinfo{year}{2024}\natexlab{}.
\newblock \showarticletitle{BEHAVIOR-1K: A Human-Centered, Embodied AI Benchmark with 1,000 Everyday Activities and Realistic Simulation}.
\newblock \bibinfo{journal}{\emph{arXiv preprint arXiv:2403.09227}} (\bibinfo{year}{2024}).
\newblock


\bibitem[Liu et~al\mbox{.}(2024)]%
        {liu2024enhancing}
\bibfield{author}{\bibinfo{person}{Jinyi Liu}, \bibinfo{person}{Yifu Yuan}, \bibinfo{person}{Jianye Hao}, \bibinfo{person}{Fei Ni}, \bibinfo{person}{Lingzhi Fu}, \bibinfo{person}{Yibin Chen}, {and} \bibinfo{person}{Yan Zheng}.} \bibinfo{year}{2024}\natexlab{}.
\newblock \showarticletitle{Enhancing robotic manipulation with AI feedback from multimodal large language models}.
\newblock \bibinfo{journal}{\emph{arXiv preprint arXiv:2402.14245}} (\bibinfo{year}{2024}).
\newblock


\bibitem[Lu et~al\mbox{.}(2022)]%
        {lu2022neuro}
\bibfield{author}{\bibinfo{person}{Yujie Lu}, \bibinfo{person}{Weixi Feng}, \bibinfo{person}{Wanrong Zhu}, \bibinfo{person}{Wenda Xu}, \bibinfo{person}{Xin~Eric Wang}, \bibinfo{person}{Miguel Eckstein}, {and} \bibinfo{person}{William~Yang Wang}.} \bibinfo{year}{2022}\natexlab{}.
\newblock \showarticletitle{Neuro-symbolic procedural planning with commonsense prompting}.
\newblock \bibinfo{journal}{\emph{arXiv preprint arXiv:2206.02928}} (\bibinfo{year}{2022}).
\newblock


\bibitem[Nair et~al\mbox{.}(2022)]%
        {nair2022r3m}
\bibfield{author}{\bibinfo{person}{Suraj Nair}, \bibinfo{person}{Aravind Rajeswaran}, \bibinfo{person}{Vikash Kumar}, \bibinfo{person}{Chelsea Finn}, {and} \bibinfo{person}{Abhinav Gupta}.} \bibinfo{year}{2022}\natexlab{}.
\newblock \showarticletitle{R3m: A universal visual representation for robot manipulation}.
\newblock \bibinfo{journal}{\emph{arXiv preprint arXiv:2203.12601}} (\bibinfo{year}{2022}).
\newblock


\bibitem[Pashevich et~al\mbox{.}(2021)]%
        {pashevich2021episodic}
\bibfield{author}{\bibinfo{person}{Alexander Pashevich}, \bibinfo{person}{Cordelia Schmid}, {and} \bibinfo{person}{Chen Sun}.} \bibinfo{year}{2021}\natexlab{}.
\newblock \showarticletitle{Episodic transformer for vision-and-language navigation}. In \bibinfo{booktitle}{\emph{Proceedings of the IEEE/CVF International Conference on Computer Vision}}. \bibinfo{pages}{15942--15952}.
\newblock


\bibitem[Sethian(1999)]%
        {sethian1999fast}
\bibfield{author}{\bibinfo{person}{James~A Sethian}.} \bibinfo{year}{1999}\natexlab{}.
\newblock \showarticletitle{Fast marching methods}.
\newblock \bibinfo{journal}{\emph{SIAM review}} \bibinfo{volume}{41}, \bibinfo{number}{2} (\bibinfo{year}{1999}), \bibinfo{pages}{199--235}.
\newblock


\bibitem[Sharan et~al\mbox{.}(2024)]%
        {sharan2024plan}
\bibfield{author}{\bibinfo{person}{SP Sharan}, \bibinfo{person}{Ruihan Zhao}, \bibinfo{person}{Zhangyang Wang}, \bibinfo{person}{Sandeep~P Chinchali}, {et~al\mbox{.}}} \bibinfo{year}{2024}\natexlab{}.
\newblock \showarticletitle{Plan Diffuser: Grounding LLM planners with diffusion models for robotic manipulation}. In \bibinfo{booktitle}{\emph{Bridging the Gap between Cognitive Science and Robot Learning in the Real World: Progresses and New Directions}}.
\newblock


\bibitem[Shridhar et~al\mbox{.}(2020a)]%
        {ALFRED20}
\bibfield{author}{\bibinfo{person}{Mohit Shridhar}, \bibinfo{person}{Jesse Thomason}, \bibinfo{person}{Daniel Gordon}, \bibinfo{person}{Yonatan Bisk}, \bibinfo{person}{Winson Han}, \bibinfo{person}{Roozbeh Mottaghi}, \bibinfo{person}{Luke Zettlemoyer}, {and} \bibinfo{person}{Dieter Fox}.} \bibinfo{year}{2020}\natexlab{a}.
\newblock \showarticletitle{{ALFRED: A benchmark for interpreting grounded instructions for everyday tasks}}. In \bibinfo{booktitle}{\emph{The IEEE Conference on Computer Vision and Pattern Recognition}}.
\newblock
\urldef\tempurl%
\url{https://arxiv.org/abs/1912.01734}
\showURL{%
\tempurl}


\bibitem[Shridhar et~al\mbox{.}(2020b)]%
        {shridhar2020alfworld}
\bibfield{author}{\bibinfo{person}{Mohit Shridhar}, \bibinfo{person}{Xingdi Yuan}, \bibinfo{person}{Marc-Alexandre C{\^o}t{\'e}}, \bibinfo{person}{Yonatan Bisk}, \bibinfo{person}{Adam Trischler}, {and} \bibinfo{person}{Matthew Hausknecht}.} \bibinfo{year}{2020}\natexlab{b}.
\newblock \showarticletitle{Alfworld: Aligning text and embodied environments for interactive learning}.
\newblock \bibinfo{journal}{\emph{arXiv preprint arXiv:2010.03768}} (\bibinfo{year}{2020}).
\newblock


\bibitem[Singh et~al\mbox{.}(2023)]%
        {singh2023progprompt}
\bibfield{author}{\bibinfo{person}{Ishika Singh}, \bibinfo{person}{Valts Blukis}, \bibinfo{person}{Arsalan Mousavian}, \bibinfo{person}{Ankit Goyal}, \bibinfo{person}{Danfei Xu}, \bibinfo{person}{Jonathan Tremblay}, \bibinfo{person}{Dieter Fox}, \bibinfo{person}{Jesse Thomason}, {and} \bibinfo{person}{Animesh Garg}.} \bibinfo{year}{2023}\natexlab{}.
\newblock \showarticletitle{Progprompt: Generating situated robot task plans using large language models}. In \bibinfo{booktitle}{\emph{IEEE International Conference on Robotics and Automation}}. IEEE, \bibinfo{pages}{11523--11530}.
\newblock


\bibitem[Skirzynski(2020)]%
        {skirzynski2020language}
\bibfield{author}{\bibinfo{person}{Julian Skirzynski}.} \bibinfo{year}{2020}\natexlab{}.
\newblock \bibinfo{booktitle}{\emph{Language-conditional imitation learning}}.
\newblock \bibinfo{publisher}{McGill University (Canada)}.
\newblock


\bibitem[Song et~al\mbox{.}(2022)]%
        {song2022one}
\bibfield{author}{\bibinfo{person}{Chan~Hee Song}, \bibinfo{person}{Jihyung Kil}, \bibinfo{person}{Tai-Yu Pan}, \bibinfo{person}{Brian~M Sadler}, \bibinfo{person}{Wei-Lun Chao}, {and} \bibinfo{person}{Yu Su}.} \bibinfo{year}{2022}\natexlab{}.
\newblock \showarticletitle{One step at a time: Long-horizon vision-and-language navigation with milestones}. In \bibinfo{booktitle}{\emph{IEEE Conference on Computer Vision and Pattern Recognition}}. \bibinfo{pages}{15482--15491}.
\newblock


\bibitem[Song et~al\mbox{.}(2023)]%
        {song2023llm}
\bibfield{author}{\bibinfo{person}{Chan~Hee Song}, \bibinfo{person}{Jiaman Wu}, \bibinfo{person}{Clayton Washington}, \bibinfo{person}{Brian~M Sadler}, \bibinfo{person}{Wei-Lun Chao}, {and} \bibinfo{person}{Yu Su}.} \bibinfo{year}{2023}\natexlab{}.
\newblock \showarticletitle{Llm-planner: Few-shot grounded planning for embodied agents with large language models}. In \bibinfo{booktitle}{\emph{IEEE International Conference on Computer Vision}}. \bibinfo{pages}{2998--3009}.
\newblock


\bibitem[Vemprala et~al\mbox{.}(2023)]%
        {vemprala2023chatgpt}
\bibfield{author}{\bibinfo{person}{Sai Vemprala}, \bibinfo{person}{Rogerio Bonatti}, \bibinfo{person}{Arthur Bucker}, {and} \bibinfo{person}{Ashish Kapoor}.} \bibinfo{year}{2023}\natexlab{}.
\newblock \bibinfo{booktitle}{\emph{ChatGPT for Robotics: design principles and model abilities}}.
\newblock \bibinfo{type}{{T}echnical {R}eport} MSR-TR-2023-8. \bibinfo{institution}{Microsoft}.
\newblock
\urldef\tempurl%
\url{https://www.microsoft.com/en-us/research/publication/chatgpt-for-robotics-design-principles-and-model-abilities/}
\showURL{%
\tempurl}


\bibitem[Wang et~al\mbox{.}(2024)]%
        {wang2024survey}
\bibfield{author}{\bibinfo{person}{Lei Wang}, \bibinfo{person}{Chen Ma}, \bibinfo{person}{Xueyang Feng}, \bibinfo{person}{Zeyu Zhang}, \bibinfo{person}{Hao Yang}, \bibinfo{person}{Jingsen Zhang}, \bibinfo{person}{Zhiyuan Chen}, \bibinfo{person}{Jiakai Tang}, \bibinfo{person}{Xu Chen}, \bibinfo{person}{Yankai Lin}, {et~al\mbox{.}}} \bibinfo{year}{2024}\natexlab{}.
\newblock \showarticletitle{A survey on large language model based autonomous agents}.
\newblock \bibinfo{journal}{\emph{Frontiers of Computer Science}} \bibinfo{volume}{18}, \bibinfo{number}{6} (\bibinfo{year}{2024}), \bibinfo{pages}{1--26}.
\newblock


\bibitem[Wang et~al\mbox{.}(2020)]%
        {wang2020minilm}
\bibfield{author}{\bibinfo{person}{Wenhui Wang}, \bibinfo{person}{Furu Wei}, \bibinfo{person}{Li Dong}, \bibinfo{person}{Hangbo Bao}, \bibinfo{person}{Nan Yang}, {and} \bibinfo{person}{Ming Zhou}.} \bibinfo{year}{2020}\natexlab{}.
\newblock \showarticletitle{Minilm: Deep self-attention distillation for task-agnostic compression of pre-trained transformers}.
\newblock \bibinfo{journal}{\emph{Advances in Neural Information Processing Systems}}  \bibinfo{volume}{33} (\bibinfo{year}{2020}), \bibinfo{pages}{5776--5788}.
\newblock


\bibitem[Xiuying et~al\mbox{.}(2022)]%
        {chen2022target}
\bibfield{author}{\bibinfo{person}{Chen Xiuying}, \bibinfo{person}{Alamro Hind}, \bibinfo{person}{Li Mingzhe}, \bibinfo{person}{Gao Shen}, \bibinfo{person}{Yan Rui}, \bibinfo{person}{Gao Xin}, {and} \bibinfo{person}{Zhang Xiangliang}.} \bibinfo{year}{2022}\natexlab{}.
\newblock \showarticletitle{Target-aware abstractive related work generation with contrastive learning}.
\newblock \bibinfo{journal}{\emph{International ACM SIGIR Conference on Research and Development in Information Retrieval}} (\bibinfo{year}{2022}).
\newblock


\bibitem[Yamada et~al\mbox{.}(2021)]%
        {yamada2021motion}
\bibfield{author}{\bibinfo{person}{Jun Yamada}, \bibinfo{person}{Youngwoon Lee}, \bibinfo{person}{Gautam Salhotra}, \bibinfo{person}{Karl Pertsch}, \bibinfo{person}{Max Pflueger}, \bibinfo{person}{Gaurav Sukhatme}, \bibinfo{person}{Joseph Lim}, {and} \bibinfo{person}{Peter Englert}.} \bibinfo{year}{2021}\natexlab{}.
\newblock \showarticletitle{Motion planner augmented reinforcement learning for robot manipulation in obstructed environments}. In \bibinfo{booktitle}{\emph{Conference on Robot Learning}}. PMLR, \bibinfo{pages}{589--603}.
\newblock


\bibitem[Yamazaki et~al\mbox{.}(2023)]%
        {Yamazaki2023narratives}
\bibfield{author}{\bibinfo{person}{Satoshi Yamazaki}, \bibinfo{person}{Jianquan Liu}, {and} \bibinfo{person}{Mohan Kankanhalli}.} \bibinfo{year}{2023}\natexlab{}.
\newblock \showarticletitle{Sequential action retrieval for generating narratives from long videos}. In \bibinfo{booktitle}{\emph{Workshop on User-Centric Narrative Summarization of Long Videos}}. \bibinfo{publisher}{Association for Computing Machinery}, \bibinfo{address}{New York, NY, USA}, \bibinfo{pages}{25–29}.
\newblock
\showISBNx{9798400702778}
\urldef\tempurl%
\url{https://doi.org/10.1145/3607540.3617143}
\showDOI{\tempurl}


\bibitem[Yu et~al\mbox{.}(2022)]%
        {yu2022bashexplainer}
\bibfield{author}{\bibinfo{person}{Chi Yu}, \bibinfo{person}{Guang Yang}, \bibinfo{person}{Xiang Chen}, \bibinfo{person}{Ke Liu}, {and} \bibinfo{person}{Yanlin Zhou}.} \bibinfo{year}{2022}\natexlab{}.
\newblock \showarticletitle{Bashexplainer: Retrieval-augmented bash code comment generation based on fine-tuned codebert}.
\newblock \bibinfo{journal}{\emph{IEEE International Conference on Software Maintenance and Evolution}}, \bibinfo{pages}{82--93}.
\newblock


\bibitem[Zare et~al\mbox{.}(2024)]%
        {zare2024rap}
\bibfield{author}{\bibinfo{person}{Ali Zare}, \bibinfo{person}{Yulei Niu}, \bibinfo{person}{Hammad Ayyubi}, {and} \bibinfo{person}{Shih-fu Chang}.} \bibinfo{year}{2024}\natexlab{}.
\newblock \showarticletitle{RAP: Retrieval-Augmented planner for adaptive procedure planning in instructional videos}.
\newblock \bibinfo{journal}{\emph{arXiv preprint arXiv:2403.18600}} (\bibinfo{year}{2024}).
\newblock


\bibitem[Zhan et~al\mbox{.}(2021)]%
        {zhan2021framework}
\bibfield{author}{\bibinfo{person}{Albert Zhan}, \bibinfo{person}{Ruihan Zhao}, \bibinfo{person}{Lerrel Pinto}, \bibinfo{person}{Pieter Abbeel}, {and} \bibinfo{person}{Michael Laskin}.} \bibinfo{year}{2021}\natexlab{}.
\newblock \showarticletitle{A framework for efficient robotic manipulation}. In \bibinfo{booktitle}{\emph{Deep RL Workshop of Neural Information Processing Systems Conference}}.
\newblock


\bibitem[Zhang et~al\mbox{.}(2023)]%
        {zhang2023syntax}
\bibfield{author}{\bibinfo{person}{Xiangyu Zhang}, \bibinfo{person}{Yu Zhou}, \bibinfo{person}{Guang Yang}, {and} \bibinfo{person}{Taolue Chen}.} \bibinfo{year}{2023}\natexlab{}.
\newblock \showarticletitle{Syntax-aware retrieval augmented code generation}.
\newblock \bibinfo{journal}{\emph{The Conference on Empirical Methods in Natural Language Processing}}.
\newblock


\bibitem[Zhang and Chai(2021)]%
        {zhang2021hierarchical}
\bibfield{author}{\bibinfo{person}{Yichi Zhang} {and} \bibinfo{person}{Joyce Chai}.} \bibinfo{year}{2021}\natexlab{}.
\newblock \showarticletitle{Hierarchical task learning from language instructions with unified transformers and self-monitoring}.
\newblock \bibinfo{journal}{\emph{arXiv preprint arXiv:2106.03427}} (\bibinfo{year}{2021}).
\newblock


\bibitem[Zhao et~al\mbox{.}(2022)]%
        {zhao2022tvln}
\bibfield{author}{\bibinfo{person}{Yusheng Zhao}, \bibinfo{person}{Jinyu Chen}, \bibinfo{person}{Chen Gao}, \bibinfo{person}{Wenguan Wang}, \bibinfo{person}{Lirong Yang}, \bibinfo{person}{Haibing Ren}, \bibinfo{person}{Huaxia Xia}, {and} \bibinfo{person}{Si Liu}.} \bibinfo{year}{2022}\natexlab{}.
\newblock \showarticletitle{Target-Driven Structured Transformer Planner for Vision-Language Navigation}. In \bibinfo{booktitle}{\emph{Proceedings of the 30th ACM International Conference on Multimedia}}. \bibinfo{publisher}{Association for Computing Machinery}, \bibinfo{address}{New York, NY, USA}, \bibinfo{pages}{4194–4203}.
\newblock
\showISBNx{9781450392037}
\urldef\tempurl%
\url{https://doi.org/10.1145/3503161.3548281}
\showDOI{\tempurl}


\bibitem[Zhuang et~al\mbox{.}(2022)]%
        {zhuang2022lvln}
\bibfield{author}{\bibinfo{person}{Yifeng Zhuang}, \bibinfo{person}{Qiang Sun}, \bibinfo{person}{Yanwei Fu}, \bibinfo{person}{Lifeng Chen}, {and} \bibinfo{person}{Xiangyang Xue}.} \bibinfo{year}{2022}\natexlab{}.
\newblock \showarticletitle{Local Slot Attention for Vision and Language Navigation}. In \bibinfo{booktitle}{\emph{Proceedings of the 2022 International Conference on Multimedia Retrieval}}. \bibinfo{publisher}{Association for Computing Machinery}, \bibinfo{address}{New York, NY, USA}, \bibinfo{pages}{545–553}.
\newblock
\showISBNx{9781450392389}
\urldef\tempurl%
\url{https://doi.org/10.1145/3512527.3531366}
\showDOI{\tempurl}


\end{thebibliography}

\end{document}